\documentclass[sigconf, authorversion]{acmart}
\usepackage{multirow}                 
\usepackage{float}                    
\usepackage{tabu}                     
\usepackage{multicol}                 
\usepackage{makecell}                 
\usepackage{booktabs}                 
\definecolor{urlcolor}{RGB}{251, 49, 153}
\AtBeginDocument{%
  }

\copyrightyear{2024}
\acmYear{2024}
\setcopyright{acmlicensed}\acmConference[MM '24]{Proceedings of the 32nd ACM International Conference on Multimedia}{October 28-November 1, 2024}{Melbourne, VIC, Australia}
\acmBooktitle{Proceedings of the 32nd ACM International Conference on Multimedia (MM '24), October 28-November 1, 2024, Melbourne, VIC, Australia}
\acmDOI{10.1145/3664647.3680756}
\acmISBN{979-8-4007-0686-8/24/10}

\begin{document}

\title{Shape-Guided Clothing Warping for Virtual Try-On}

\author{Xiaoyu Han}
\email{xyhan@stu.hit.edu.cn}
\affiliation{%
  \institution{Harbin Institute of Technology}
  \city{Weihai}  
  \country{China}
}

\author{Shunyuan Zheng}
\email{sawyer0503@hit.edu.cn}
\affiliation{%
  \institution{Harbin Institute of Technology}
  \city{Weihai}  
  \country{China}
}

\author{Zonglin Li}
\email{zonglin.li@hit.edu.cn}
\affiliation{%
  \institution{Harbin Institute of Technology}
  \city{Weihai}  
  \country{China}
}

\author{Chenyang Wang}
\email{c.wang@stu.hit.edu.cn}
\affiliation{%
  \institution{Harbin Institute of Technology}
  \city{Weihai}  
  \country{China}
}

\author{Xin Sun}
\email{sunxintyc@hit.edu.cn}
\affiliation{%
  \institution{Harbin Institute of Technology}
  \city{Weihai}  
  \country{China}
}

\author{Quanling Meng}
\email{quanling.meng@hit.edu.cn}
\authornote{Corresponding author.}
\affiliation{%
  \institution{Harbin Institute of Technology}
  \city{Weihai}  
  \country{China}
}

\renewcommand{\shortauthors}{Xiaoyu Han et al.}

\begin{abstract}
Image-based virtual try-on aims to seamlessly fit in-shop clothing to a person image while maintaining pose consistency. Existing methods commonly employ the thin plate spline (TPS) transformation or appearance flow to deform in-shop clothing for aligning with the person's body. Despite their promising performance, these methods often lack precise control over fine details, leading to inconsistencies in shape between clothing and the person's body as well as distortions in exposed limb regions. To tackle these challenges, we propose a novel shape-guided clothing warping method for virtual try-on, dubbed SCW-VTON, which incorporates global shape constraints and additional limb textures to enhance the realism and consistency of the warped clothing and try-on results. To integrate global shape constraints for clothing warping, we devise a dual-path clothing warping module comprising a shape path and a flow path. The former path captures the clothing shape aligned with the person's body, while the latter path leverages the mapping between the pre- and post-deformation of the clothing shape to guide the estimation of appearance flow. Furthermore, to alleviate distortions in limb regions of try-on results, we integrate detailed limb guidance by developing a limb reconstruction network based on masked image modeling. Through the utilization of SCW-VTON, we are able to generate try-on results with enhanced clothing shape consistency and precise control over details. Extensive experiments demonstrate the superiority of our approach over state-of-the-art methods both qualitatively and quantitatively.
The code is available at \textcolor{urlcolor}{\url{https://github.com/xyhanHIT/SCW-VTON}}.
\end{abstract}


\begin{teaserfigure}
\begin{center}
    \vspace{-5pt}
    \includegraphics[width=1\textwidth]{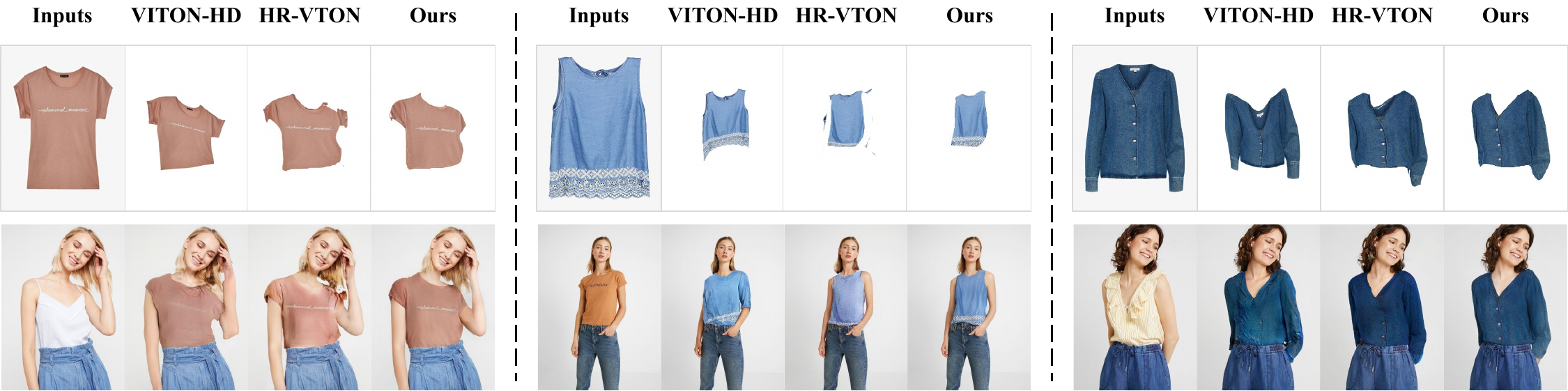}
    \vspace{-15pt}
    \captionof{figure}{Virtual try-on typically deforms clothing to fit the person's body, which is then combined with the person image to obtain try-on results. Compared with existing TPS-based methods (e.g., VITON-HD~\cite{viton-hd}) and appearance flow based methods (e.g., HR-VTON~\cite{hr-vton}), our method excels in capturing the clothing shape conforming to the person's body.}
    \label{fig:teaser} 
    \vspace{5pt}
\end{center}
\end{teaserfigure}


\maketitle

\section{Introduction}
In recent years, the e-commerce industry has experienced rapid development, leading to an increasing number of consumers purchasing clothing online. 
To provide online customers with a shopping experience that rivals in-store try-on, significant attention has been devoted to exploring the virtual try-on technology. 
Virtual try-on can be broadly categorized into 3D-based~\cite{patron, clothcap, drape, design-garment, transfer-texture, collision-handling}  and image-based methods~\cite{viton, cp-vton, acgpn, pf-afn, clothflow, pl-vton2, gp-vton, flow-style-vton, rt-vton, dgp, pasta-gan, dcton, kgi, m2e, wuton, vtnfp, fashionon, mv-vton, sc-vton, was-vton}, with the latter garnering more interest due to its lightweight data and wider applicability. This study will specifically delve into image-based virtual try-on.

Among the key processing steps in image-based virtual try-on, clothing warping stands out as particularly challenging. Its objective is to deform the target clothing to align with the person's body, which is vital for maintaining consistency between the clothing shape and the person's posture. 
Early methods~\cite{viton, cp-vton, cp-vton+, acgpn} adopt the thin plate spline (TPS) transformation~\cite{tps} to achieve this deformation. 
However, these TPS-based methods, constrained by limited degrees of freedom, prove inadequate when significant geometric deformations are needed~\cite{z-flow, clothflow}, such as the misaligned clothing generated by VITON-HD~\cite{viton-hd} in Figure \ref{fig:teaser}. 
To overcome this limitation, recent studies~\cite{clothflow,z-flow,sdafn,rmgn,sal-vton} introduce appearance flow~\cite{af} as a solution, which provides dense pixel-level predictions for deforming from the source clothing to the target one, thereby enhancing the accuracy of the resulting warped clothing. 
While these methods have demonstrated promising performance in virtual try-on, a notable limitation remains in their current inability to exert precise control over the details of the obtained warped clothing.
Firstly, the lack of the appropriate constraint on shape during pixel-level prediction may lead to inconsistencies in the shape of the warped clothing compared to the person's body. For instance, as shown in the HR-VTON~\cite{hr-vton} column of Figure \ref{fig:teaser}, the generated clothing edges appear torn and inconsistent.
Secondly, distortions may occur partially in the try-on result, especially in the limb regions not covered by clothing. This is attributed to the absence of detailed information guidance for these parts, as existing methods often use the person image with masked clothing and limb regions as the input for try-on synthesis. 

In this study, we direct our attention towards addressing these underexplored challenges. 
We present SCW-VTON, a shape-guided clothing warping method for virtual try-on, which incorporates global shape constraints and additional limb texture references to enhance the realism of generated warped clothing and try-on results. 
Initially, we leverage global shape constraints to facilitate the clothing warping phase, alleviating discrepancies between the warped clothing shape and the person’s body. 
We design a dual-path clothing warping module comprising a shape path and a flow path. 
The shape path first predicts the shape of the target clothing, facilitating the subsequent exploration of the mapping between the pre- and post-deformation clothing shapes using a set of shape-guided cross-attention blocks. 
The acquired mapping is then integrated into a flow path as global shape constraints to steer the estimation of appearance flow. This flow is subsequently applied to in-shop clothing to create the target warped clothing. 
It is worth mentioning that introducing additional global shape constraints frees our method from reliance on the original input shape, which implies that our method is also capable of transferring textures from a shapeless texture map to the person image while preserving their distribution.
Furthermore, we employ a co-training strategy for the two paths during training, which positively affects the weight update of the flow estimation model and further enhances the stability and accuracy of appearance flow.
In the subsequent process of try-on synthesis, to alleviate distortions in limb regions, we develop an additional limb reconstruction network based on the masked image modeling method. 
Specifically, we first derive a limb map from the source person image. Then, we employ an autoencoder to take the masked partial limb component as input to learn latent limb representations, from which we reconstruct realistic limb textures at specific locations. 
Through the combination of shape-guided appearance flow and reconstructed limb textures, our approach can ultimately produce try-on results with enhanced clothing shape consistency and precise control over details. 

The contributions of this paper are summarized as follows:
\begin{itemize}
    \item
    We introduce a shape-guided clothing warping method for virtual try-on that incorporates global shape constraints on clothing deformation, resulting in realistic warped clothing that conforms accurately to the person's body.
    \item
    We design a limb reconstruction network to provide precise guidance on generating limb regions of the try-on result, effectively addressing issues such as performance degradation and distortions during the try-on synthesis process.
    \item
    Extensive experiments demonstrate the superior performance of our proposed SCW-VTON compared to existing state-of-the-art methods for virtual try-on.
\end{itemize}

\section{Related Work}
\subsection{Appearance Flow}
Appearance flow is applied to predict a 2D vector field and warp the source image to the target based on the similarity in appearance, which is proposed by Zhou et al.~\cite{af} to solve the problem of novel view synthesis.
In recent years, appearance flow has been widely applied in various fields of computer vision.
In image inpainting tasks~\cite{flow-inpaint1, flow-inpaint2}, appearance flow is used to propagate pixels from source to missing regions, enhancing realism in generated contents. 
Additionally, appearance flow also enables human pose transfer~\cite{pose_transfer1, pose_transfer2, pose_transfer3, pose_transfer4}, synthesizing novel poses by warping the feature representations of the human body. 
Besides, virtual try-on has attracted widespread attention in recent years and many methods~\cite{pf-afn, pl-vton, clothflow, sdafn} introduce the appearance flow to deform the in-shop clothing to achieve the alignment with the person's body.

\begin{figure*}[!t]
    \includegraphics[width=\textwidth]{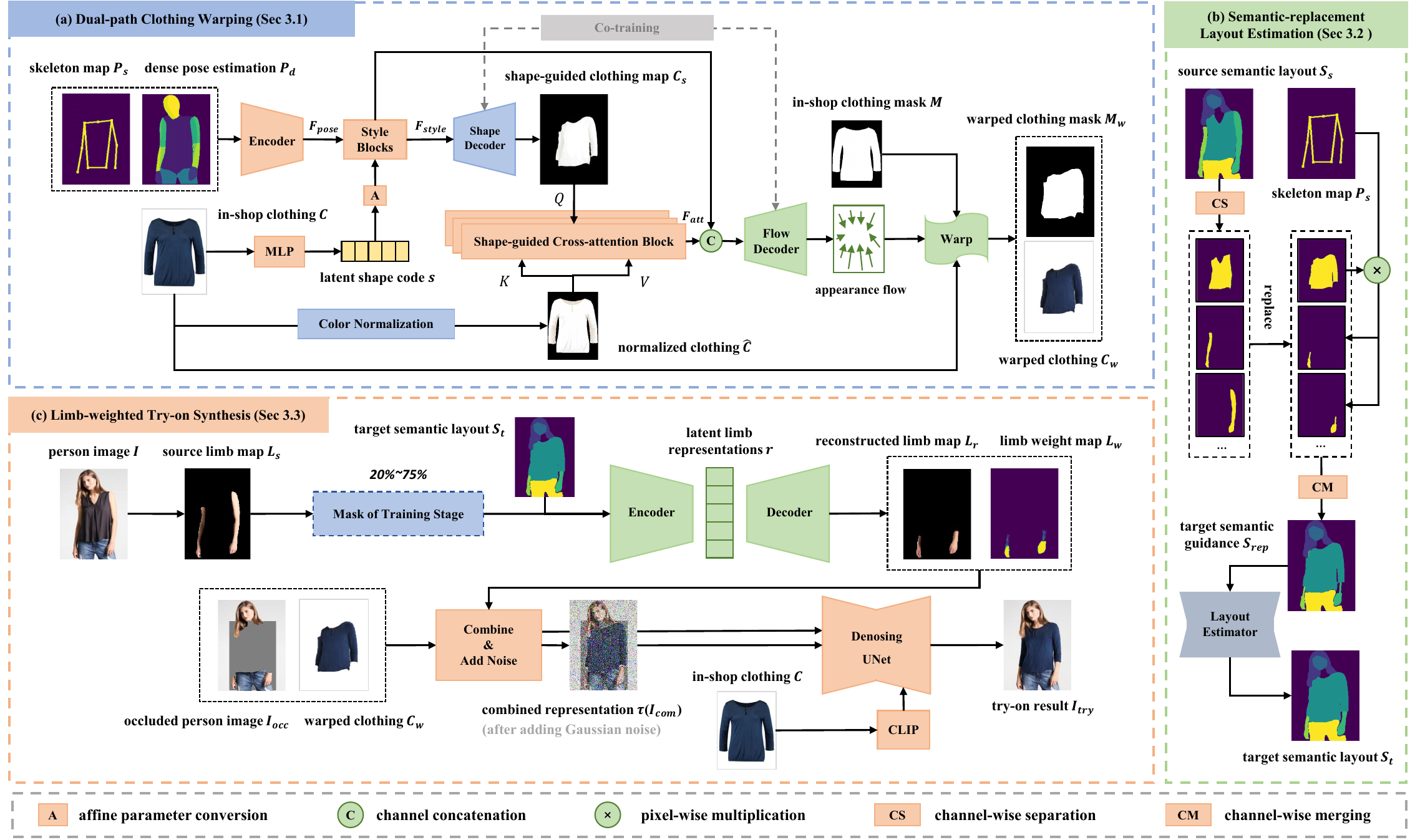}
    \vspace{-12pt}
   \caption{The overview of SCW-VTON. (a) Warped clothing $C_{w}$ is produced through a dual-path architecture, where the cross-attention features $F_{att}$ are created as global shape constraints to assist in estimating appearance flow. (b) Based on a semantic-replacement strategy, a target semantic layout $S_{t}$ that describes the person wearing target clothing is predicted. (c) A pre-trained autoencoder produces the reconstructed limb map $L_r$ from latent limb representations $r$, which is then combined with the occluded person image $I_{occ}$ and warped clothing $C_w$ to produce the final try-on result $I_{try}$ with a diffusion model.}
   \vspace{-2mm}
\label{fig:overview}
\end{figure*}

\subsection{Image-based Virtual Try-on}
\noindent\textbf{Clothing Warping.}
Clothing Warping is typically a key processing step of most image-based virtual try-on methods. 
Early methods~\cite{viton, cp-vton, cp-vton+, acgpn} primarily achieve this through thin-plate spline (TPS) transformation.
However, TPS-based methods are unable to accurately handle large geometric deformation due to the limited degrees of freedom.
Therefore, ClothFlow~\cite{clothflow} proposes an appearance flow based method to perform clothing warping with pixel-level displacements. 
Then, Zflow~\cite{z-flow} adopts gated appearance flow to further stabilize the deformation.
SAL-VTON~\cite{sal-vton} links the clothing with the person's body with semantically associated landmarks and proposes the local flow to alleviate the misalignment.
However, considering the potential drastic deformation caused by appearance flow, we introduce extra global shape constraints based on a cross-attention mechanism, which warrants a more stable flow and makes the warped clothing match a consistent shape.

\noindent\textbf{Try-on Synthesis.}
Another important step is try-on synthesis, which refers to synthesizing the target clothing and the person image to produce the try-on result.
Benefiting from the powerful generation capability of the diffusion model~\cite{denoising, diffusion}, some existing methods~\cite{dci-vton, ladi-vton, stableviton, tryondiffusion, MGD} introduce it as the final component of try-on synthesis.
DCI-VTON~\cite{dci-vton} treats virtual try-on as an inpainting task and incorporates the diffusion model to refine a coarse result.
LaDI-VTON~\cite{ladi-vton} employs textual inversion technique in virtual try-on and proposes EMASC modules to enhance synthesis quality.
StableVITON~\cite{stableviton} proposes a zero cross-attention block used in the pre-trained diffusion decoder to learn the semantic correspondence between the clothing and the person for achieving better detail preservation.
However, most methods require the synthesis network to reconstruct limb textures in the absence of corresponding reference cues from the person image, leading to the degradation of performance and distortions in try-on results. We attempt to alleviate this problem by creating an additional limb reconstruction network based on masked image modeling.

\section{Proposed Method}
As shown in Figure \ref{fig:overview}, SCW-VTON consists of three modules: 1) a dual-path clothing warping module, 2) a semantic-replacement layout estimation module, and 3) a limb-weighted try-on synthesis module. The dual-path clothing warping module produces the warped clothing $C_w$ and the warped clothing mask $M_w$ while the semantic-replacement layout estimation module generates the target semantic layout $S_{t}$. Based on these acquired results, the limb-weighted try-on synthesis module ultimately generates the try-on result $I_{try}$ through a diffusion model.

\subsection{Dual-path Clothing Warping}
\label{module1}
Figure \ref{fig:overview} (a) shows the schematic of the dual-path clothing warping module, which can be divided into a shape path, shape-guided cross-attention blocks, and a flow path.
We begin by introducing the shape characteristics, which will be used in the shape path and shape-guided cross-attention blocks.

\begin{figure}[!t]
    \includegraphics[width=1.0\linewidth]{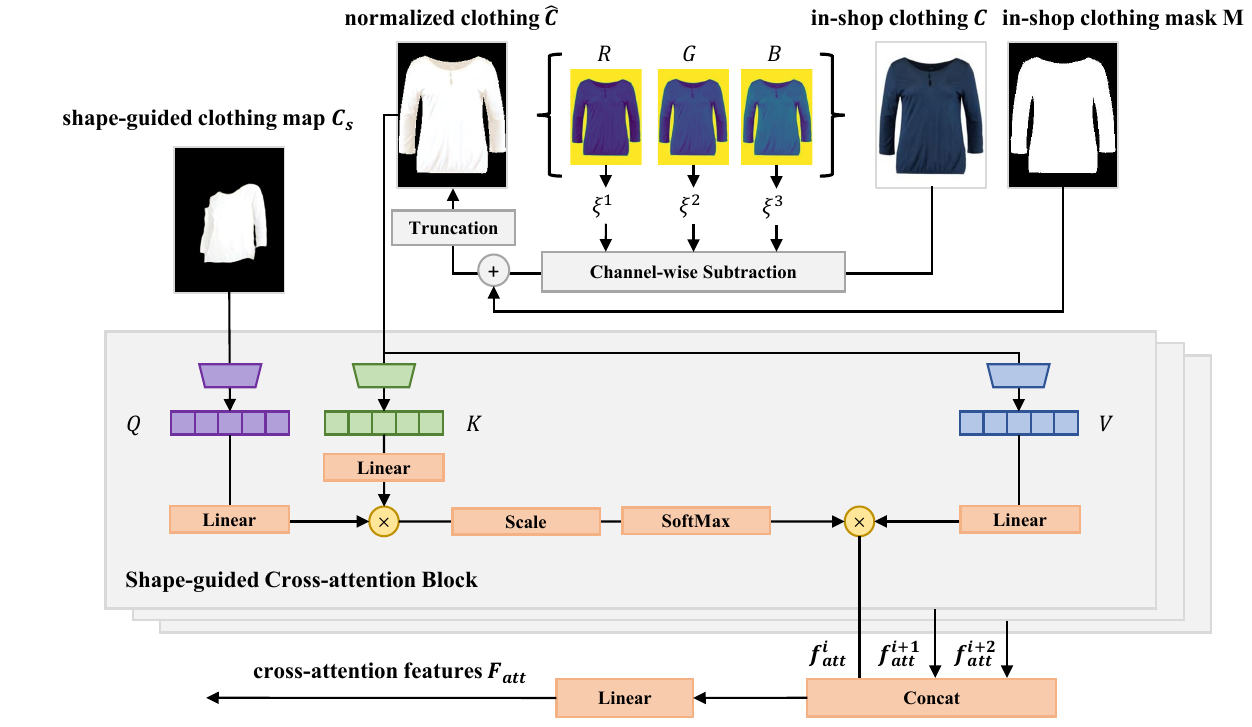}
   \caption{Shape-guided cross-attention blocks of SCW-VTON.}
\label{fig:stage1}
\end{figure}

\noindent\textbf{Shape Characteristics.} 
The first consideration is how to obtain the shape characteristics of clothing pre- and post-deformation, as this is essential for constructing the mapping between them.
In this section, we design a color normalization strategy to obtain the shape characteristics of clothing before deformation firstly, which explicitly normalizes each color channel of in-shop clothing $C$ to discard its raw color while keeping the overall gradient difference.
Specifically, as shown in Figure \ref{fig:stage1}, we first calculate the average color value $\xi^{k}$ for each channel $k$ of in-shop clothing $C$. 
Then we subtract $\xi^{k}$ from each pixel value within the clothing regions of $C$. 
Finally, we add in-shop clothing mask $M$ to the subtracted result and get the normalized clothing $\hat{C}$, which can reflect the shape characteristics of in-shop clothing more intuitively while excluding interfering information. The above process can be expressed as:
\begin{equation}
    \xi^{k} = \frac{\sum^{H_C}_{i=0}\sum^{W_C}_{j=0}(C^{k,i,j}\odot M^{i,j})}{\sum^{H_C}_{i=0}\sum^{W_C}_{j=0}M^{i,j}},
\end{equation}
\begin{equation}
    \hat{C}^{k} = \mathcal{T}(C^{k} - \xi^{k} + M),
\label{eq.1}
\end{equation}
where $\odot$ is the element-wise multiplication, $i$ and $j$ is the position of a sample pixel, $H_C$ and $W_C$ are the height and width of $C$, respectively. $\mathcal{T}(\cdot)$ is a truncated function ensuring the output value falls within the range of zero and one.

\begin{figure}[!t]
    \includegraphics[width=0.85\linewidth]{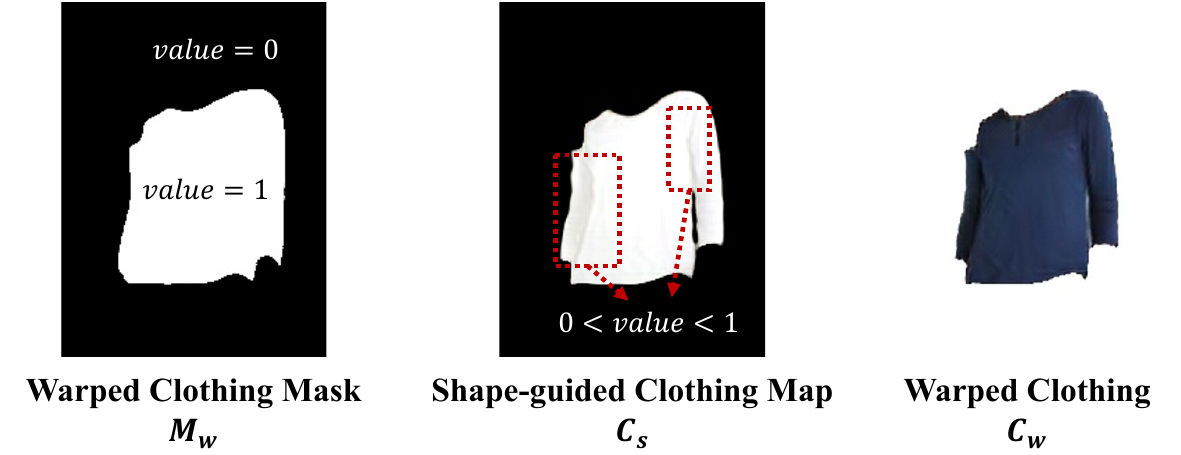}
    \vspace{-2mm}
   \caption{Compared to the binary mask $M_w$, our shape-guided clothing map $C_s$ contains detailed shape characteristics of clothing such as the basic appearance, shadows, and wrinkles.}
   \vspace{-2mm}
\label{fig:detail}
\end{figure}

\noindent\textbf{Shape Path.}
The subsequent goal is to obtain the shape characteristics of clothing after deformation, which are then used to match the normalized clothing $\hat{C}$.
We first adopt a CNN-based encoder to extract human pose features $F_{pose}$ from the skeleton map $P_s$ and dense pose estimation $P_d$, where $P_s$ is obtained by connecting 9 key points of the person's body and $P_d$ is from \cite{densepose}. 
Although $F_{pose}$ already contains basic information that matches the person's body, it is also essential to ensure that the design and style of the generated clothing are consistent with the in-shop clothing $C$.
Inspired by \cite{style_transfer}, we adopt a multilayer perceptron (MLP) to embed in-shop clothing $C$ into the latent space as a global style code $s$, which is used to calculate a set of affine transformation parameters in style blocks to adjust the pose features $F_{pose}$ and get the stylized shared features $F_{style}$ for the shape decoder and the subsequent flow decoder. This process can be formulated as:
\begin{equation}
    F^{k,i,j}_{style} = \gamma^{k,i,j}(s)\frac{F^{k,i,j}_{pose}-\mu^{k}}{\sigma^{k}} + \delta^{k,i,j}(s),
\end{equation}
where $F^{k,i,j}_{style}$ is a particular sample of $F_{style}$ at location $(k,i,j)$, $\gamma(\cdot)$ and $\delta(\cdot)$ are the convolution operations that convert the input into affine parameters, $\mu^{k}$ and $\sigma^{k}$ are the mean and standard deviation of pose features $F^{k}_{pose}$, respectively. $\mu^k$ is calculated as:
\begin{equation}
    \mu^{k} = \frac{1}{H_{F}W_{F}}\sum^{H_{F}}_{i=0}\sum^{W_{F}}_{j=0}F^{k,i,j}_{pose},
\end{equation}
where $H_{F}$ and $W_{F}$ are the height and width of $F_{pose}$, respectively. $\sigma^k$ is calculated as:
\begin{equation}
    \sigma^{k} = \sqrt{\frac{1}{H_{F}W_{F}}\sum^{H_{F}}_{i=0}\sum^{W_{F}}_{j=0}(F^{k,i,j}_{pose}-\mu^{k})^{2}}.
\end{equation}
Then, we adopt a shape decoder to up-sample the shared features $F_{style}$ and generate a shape-guided clothing map $C_s$, where we employ our color normalization strategy again to get the ground truth of $C_s$ in the training phase (please refer to Section \ref{loss} for more details). Note that $C_s$ is not the binary mask of the warped clothing, as shown in Figure \ref{fig:detail}, compared to the binary mask $M_w$ obtained by directly applying flow to the in-shop clothing mask $M$, $C_s$ encompasses detailed shape characteristics of clothing after deformation such as the basic appearance, shadows, and wrinkles.

\noindent\textbf{Shape-guided Cross-attention Blocks.}
After obtaining the normalized clothing $\hat{C}$ and the shape-guided clothing map $C_s$ that respectively reflect the shape characteristics of clothing before and after deformation, we adopt cross-attention blocks to explore the mapping between them, which is subsequently used to constrain the estimation of appearance flow.
As shown in Figure \ref{fig:stage1}, $C_s$ and $\hat{C}$ are fed into three separate encoders to get $Q$,$K$,$V$, respectively:
\begin{equation}
    Q = Fla(E_q(C_s)),K = Fla(E_k(\hat{C})),V = Fla(E_v(\hat{C})),
\end{equation}
where $E_q$, $E_k$, $E_v$ are encoders to get the corresponding components of the cross-attention mechanism, and $Fla(\cdot)$ is the flattening operation for extracted feature maps.
Next, for the $i$-th cross-attention block, its output can be represented as:
\begin{equation}
    f^{i}_{att} = softmax(\frac{QA^i_{q}(KA^i_{k})^T}{\sqrt{d}})VA^i_v,
\end{equation}
where $A_q$, $A_k$, $A_v$ represent linear layers, and $d$ is the dimension of $KA^i_k$. Finally, all the outputs of cross-attention blocks are integrated to form the cross-attention features $F_{att}$.

\noindent\textbf{Flow Path.}
The essence of appearance flow is the pixel-wise displacement, which does not involve the degradation of image quality caused by down-sampling, so a warped result applied by appearance flow has the ability to retain detailed textures. However, directly estimating appearance flow without global shape constraints may lead to unnatural drastic deformations.
Therefore, instead of directly decoding the shared features $F_{style}$ to get the appearance flow, we introduce $F_{att}$ as extra global shape constraints to perform the shape-guided warping.
Specifically, the inputs to the flow decoder consist of two parts: one is the shared features $F_{style}$, and the other is the cross-attention features $F_{att}$. They are concatenated and fed into the flow encoder together to perform pixel-wise regression and estimate the appearance flow. 
Note that at the training stage, the weight parameters of the shape decoder and the flow decoder are updated simultaneously. This enables the flow decoder to collaborate with the shape decoder seamlessly and implicitly, thereby predicting the appearance flow with a more consistent shape.
Finally, the appearance flow is applied to the in-shop clothing $C$ and the in-shop clothing mask $M$, getting the warped clothing $C_w$ and the corresponding warped clothing mask $M_{w}$.

\subsection{Semantic-replacement Layout Estimation}
\label{module2}
Given the warped clothing mask $M_{w}$, the skeleton map $P_s$, and the source semantic layout $S_{s}$, the goal of the semantic-replacement layout estimation module is to produce the target semantic layout $S_{t}$, which describes the person wearing new clothing. $S_{t}$ is utilized to provide the location information for limb reconstruction in the following process of try-on synthesis.

Since it is almost unavailable to acquire the training data about a person wearing two different clothing in a fixed pose, it is a necessary and common practice to acquire person representations that discard the person's original clothing information to train the network in a self-supervised way.
We use a semantic-replacement strategy to perform the above discarding during predicting $S_t$.
As shown in Figure \ref{fig:overview} (b), the source semantic layout $S_{s}$ is separated into a multi-channel binary parsing map and each channel corresponds to clothing or a part of the person's body.
We utilize the warped clothing mask $M_{w}$ and a portion of the skeleton map $P_s$ to replace the original clothing and limb channels to get the target semantic guidance $S_{rep}$, which ensures semantic continuity and pose consistency while preventing negative interference with the network training caused by the person's original clothing. 
Subsequently, we use a UNet~\cite{unet} model as the semantic layout estimator, which takes $S_{rep}$ as the input and predicts the target semantic layout $S_{t}$.
More detailed descriptions of this module can be found in the supplementary material.

\begin{figure}[!t]
    \includegraphics[width=1.0\linewidth]{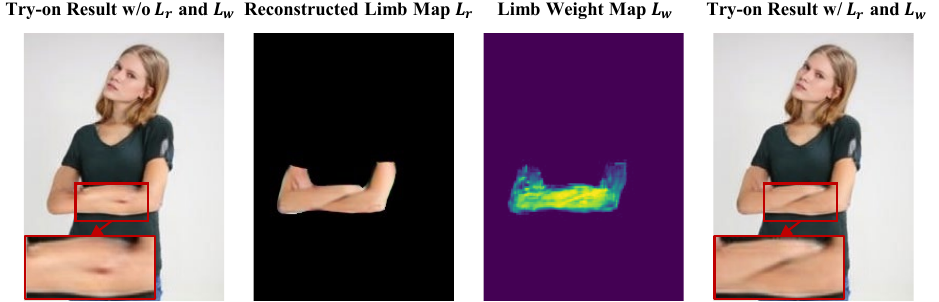}
   \caption{Comparsion of the try-on result w/o and w/ the reconstructed limb map $L_r$ and the limb weight map $L_w$.}
   \vspace{-12pt}
\label{fig:limb}
\end{figure}

\begin{figure*}[!t]
    \centering
    \includegraphics[width=1.0\textwidth]{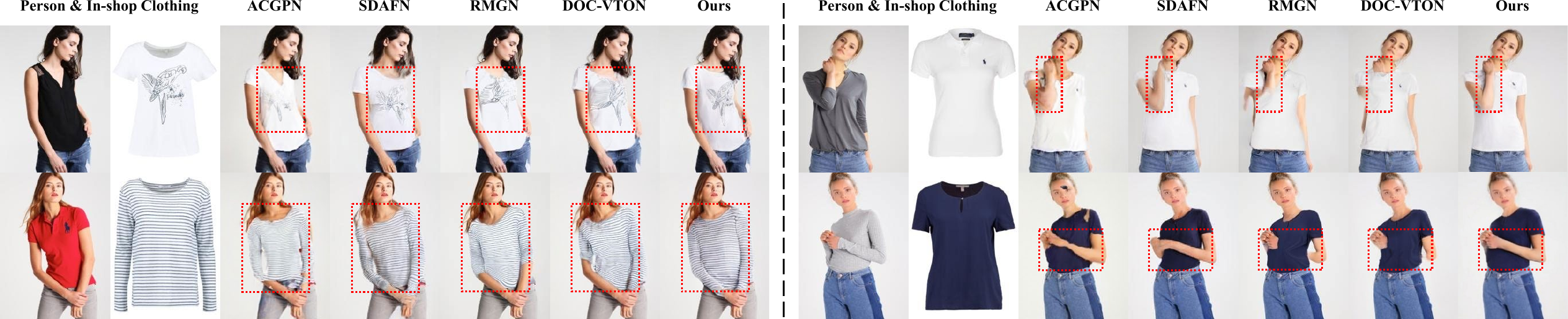}
    \vspace{-15pt}
    \caption{Qualitative results of baseline methods and our SCW-VTON on the VITON~\cite{viton} dataset.}
    \vspace{-5pt}
\label{fig:quality}
\end{figure*}

\subsection{Limb-weighted Try-on Synthesis}
\label{module3}
After obtaining the warped clothing $C_{w}$ and the target semantic layout $S_{t}$, our final goal is to generate the try-on result.
However, during this process, reconstructing limb textures is a challenging task, as the lack of sufficient reference information may lead to inconsistency between the distribution of the reconstructed result and the original image.

\noindent\textbf{Limb Reconstruction.} Therefore, we propose a separate limb reconstruction network, which focuses on learning compact latent representations of the person's limbs through masked image modeling, aiming to enhance the realism and consistency of the reconstructed limb textures with the original data.
As depicted in Figure \ref{fig:overview} (c), we first extract the source limb map $L_s$ from the person image $I$.
Inspired by \cite{mae}, we randomly mask $L_s$ by 20\%$\sim$75\% and obtain the masked limb map before an autoencoder.
On the one hand, data pairs with the asymmetric input and output are created in the training stage, enhancing the ability of the autoencoder to reconstruct limb textures. On the other hand, random masking encourages the network to learn more compact limb representations.
Note that the mask operation is only set in the training phase, while during testing, the autoencoder directly takes $L_s$ as input.
Besides, the target semantic layout $S_t$ is another input of the autoencoder, which is used to provide location information for limb reconstruction.
The autoencoder is pre-trained individually and is capable of outputting the reconstructed limb map $L_r$ and the limb weight map $L_w$, where the latter is a correlation weight to adaptively modify the limb regions that may be prone to performance degradation and distortions (as shown in Figure \ref{fig:limb}).

\noindent\textbf{Try-on Synthesis.} Based on $L_w$, $L_r$ is then combined with the occluded person image $I_{occ}$ (obtained by occluding the clothing and limb regions of $I$) and the warped clothing $C_w$ to get the combined representation $I_{com}$.
Inspired by \cite{dci-vton}, we adopt the pre-trained diffusion model as the final try-on synthesis network, utilizing the combined representation $I_{com}$ as the input to produce the try-on result $I_{try}$. 
Besides, a pre-trained CLIP~\cite{clip} image encoder is employed to extract additional conditions, which are injected into the Denoising UNet through the cross-attention mechanism to guide the generation of the diffusion model.

\subsection{Training Objectives}
\label{loss}
\noindent\textbf{Dual-path Clothing Warping.} To train the dual-path clothing warping module, we first adopt the reconstruction loss to constrain the pixel-wise value of $C_{s}$ and $C_{w}$, which is formulated as:
\begin{equation}
    l_{rec} = \Vert C_{s} - \hat{C}_{gt}\Vert_{1} + \Vert C_{w} - C_{gt}\Vert_{1},
\end{equation}
where $C_{gt}$ is the ground truth of the warped clothing $C_w$, it is extracted from the person image $I$. $\hat{C}_{gt}$ is obtained by applying the color normalization strategy to $C_{gt}$.
Besides, we adopt the perceptual loss proposed in \cite{vgg_loss} to calculate the distance of the features extracted by the VGG-19~\cite{vgg19} network:
\begin{equation}
    l_{per} = \sum^5_{i=1} ( \Vert \phi_i(C_{s}) - \phi_i(\hat{C}_{gt})\Vert_1 + \Vert \phi_i(C_{w}) - \phi_i(C_{gt})\Vert_1 ),
\end{equation}
where $\phi_i(\cdot)$ denotes the feature maps of the $i$-th layer in the pretrained perception network.
Furthermore, we adopt a mask loss to constrain the warped clothing mask $M_w$, which is formulated as:
\begin{equation}
    l_{mask} = \Vert M_{w} - M_{gt}\Vert_{1},
\end{equation}
where $M_{gt}$ is the mask of $C_{gt}$.
Overall, the loss of the dual-path clothing warping module is represented as:
\begin{equation}
    l_{warp} = l_{rec} + \lambda_{per} l_{per} + l_{mask},
\end{equation}
where $\lambda_{per}$ is used to balance the weights of these losses.

\noindent\textbf{Semantic-replacement Layout Estimation.} 
We use a weighted cross-entropy loss to supervise the training process of the semantic layout estimator, which is expressed as:
\begin{equation}
    l_{sem} =-\frac{1}{n}\sum^{n}_{i=0}\sum^{c}_{j=0}{w_{j}}{S^{i,j}_{s}}log(S^{i,j}_{t}),
\end{equation}
where $n$ denotes the number of samples, $c$ is the number of channels of $S_{s}$ and $S_{t}$, and $w_j$ is the loss weight in the $j$-th class channel.

\noindent\textbf{Limb-weighted Try-on Synthesis.} Finally, following \cite{dci-vton}, the loss of the limb-weighted try-on synthesis $l_{syn}$ is designed in two parts, which can be represented as:
\begin{equation}
    \begin{split}
        l_{syn} = \lambda_{vgg}l_{vgg} + l_{ldm}.
    \end{split}    
\end{equation}
$l_{vgg}$ is similar to $l_{per}$:
\begin{equation}
 \begin{split}
    l_{vgg} = &\sum^5_{i=1} \Vert \phi_i(I_{try}) - \phi_i(I)\Vert_1,
 \end{split}
\end{equation}
and $l_{ldm}$ is formulated as:
\begin{equation}
    \begin{split}
        l_{ldm} = \Vert \epsilon-\epsilon_\theta(\tau(\mathcal{E}(I_{com})), \mathcal{E}(I_{com}), m, CLIP(C), t)\Vert^{2}_{2},
    \end{split}    
\end{equation}
where $\mathcal{E}$ is a pre-trained encoder belonging to the diffusion model, which embeds the images from image space to latent space. $\tau(\cdot)$ is the operation of adding noise, $m$ is the mask used to occlude the person image $I$, and $t$ is the timestamp.

\begin{figure*}[!t]
    \centering
    \includegraphics[width=1.0\textwidth]{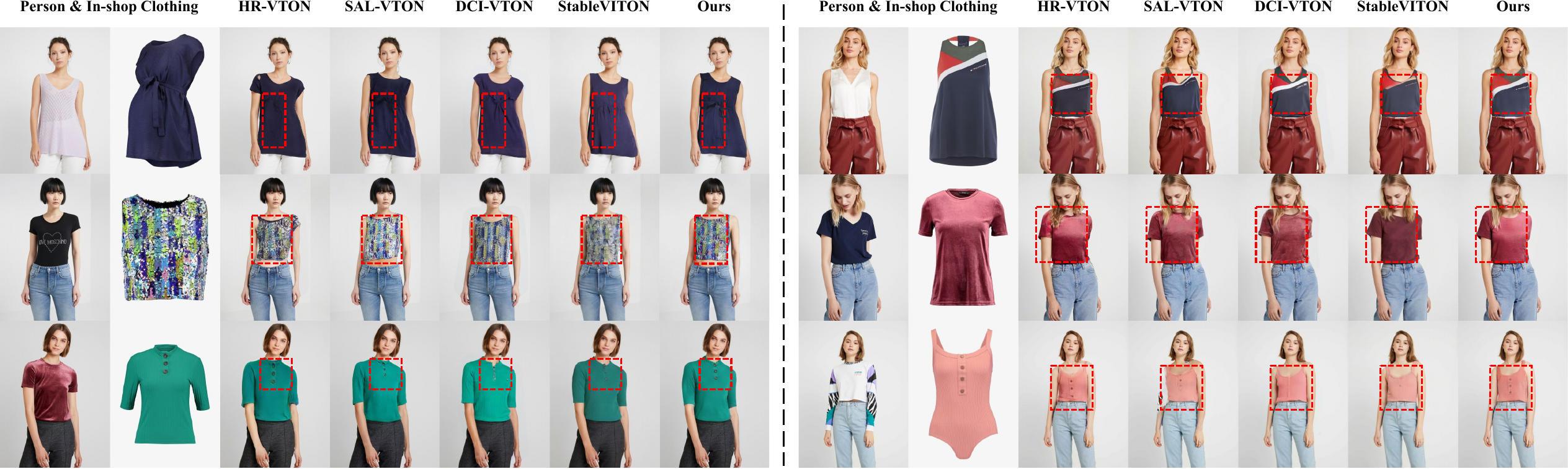}
    \vspace{-15pt}
   \caption{Qualitative results generated by baseline methods and our SCW-VTON on the VITON-HD~\cite{viton-hd} dataset.}
   \vspace{-2pt}
\label{fig:quality-hd}
\end{figure*}

\begin{table*}[!t]\small
\centering
\caption{Quantitative comparisons on the VITON~\cite{viton} and VITON-HD~\cite{viton-hd} datasets. The percentage results in the last column are displayed as a/b, with a and b representing the preference for the baseline method and our method, respectively.}
\vspace{-3pt}
\setlength{\tabcolsep}{12pt}{
\scalebox{1}{
\begin{tabular}{l|c|cc|ccc|c}
\Xhline{2\arrayrulewidth}
\multirow{2}{*}{Method}                      & \multirow{2}{*}{Dataset}        & \multicolumn{2}{c|}{Warped Clothing} & \multicolumn{3}{c|}{Try-on Results}                  & \multirow{2}{*}{User Study}  \\ \cline{3-7} 
\multicolumn{1}{c|}{}                        &                                 & SSIM (↑)          & PSNR (↑)         & FID (↓)        & SSIM (↑)          & PSNR (↑)        &                        \\ \hline \hline
ACGPN~\cite{acgpn}                           & \multirow{6}{*}{VITON}          & 0.8711            & 22.38            & 12.77          & 0.8454            & 23.12           & 12.83\% / 87.17\%       \\
PL-VTON~\cite{pl-vton}                       &                                 & 0.8434            & 19.94            & 11.68          & 0.8507            & 24.73           & 18.67\% / 81.33\%       \\
SDAFN~\cite{sdafn}                           &                                 & -                 & -                & 10.83          & 0.8399            & 23.49           & 31.33\% / 68.67\%        \\
RMGN~\cite{rmgn}                             &                                 & 0.8588            & 21.28            & 10.52          & 0.8633            & 24.95           & 29.83\% / 70.17\%        \\ 
DOC-VTON~\cite{OccluMix}                     &                                 & 0.8531            & 19.05            & 9.43           & 0.8323            & 22.07           & 33.02\% / 66.98\%        \\
\textbf{SCW-VTON (ours)}                     &                                 & \textbf{0.8839}   & \textbf{23.31}   & \textbf{8.89}  & \textbf{0.8897}   & \textbf{26.78}  &   references              \\ \hline
VITON-HD~\cite{viton-hd}                     & \multirow{8}{*}{VITON-HD}       & 0.8641            & 18.29            & 14.05          & 0.8497            & 21.19           & 23.08\% / 76.92\%         \\ 
HR-VTON~\cite{hr-vton}                       &                                 & 0.8622            & 18.41            & 11.69          & 0.8657            & 22.38           & 29.28\% / 70.72\%         \\
SAL-VTON~\cite{sal-vton}                     &                                 & -                 & -                & 9.64           & 0.8798            & 23.29           & 36.72\% / 63.28\%         \\
GP-VTON~\cite{gp-vton}                       &                                 & 0.8905            & 21.76            & 9.52           & 0.8735            & 23.31           & 36.08\% / 63.92\%         \\
LaDI-VTON~\cite{ladi-vton}                   &                                 & 0.8897            & 21.88            & 9.57           & 0.8638            & 22.52           & 31.80\% / 68.20\%          \\
DCI-VTON~\cite{dci-vton}                     &                                 & 0.8841            & 21.15            & 9.67           & 0.8749            & 23.07           & 35.20\% / 64.80\%         \\
StableVITON~\cite{stableviton}               &                                 & -                 & -                & 9.45           & 0.8678            & 23.48           & 39.32\% / 60.68\%        \\
\textbf{SCW-VTON (ours)}                     &                                 & \textbf{0.8971}   & \textbf{22.47}   & \textbf{8.96}  & \textbf{0.8829}   & \textbf{23.98}  &  references                \\ 
\Xhline{2\arrayrulewidth}
\end{tabular}}
\label{tab:quantitative}}
\end{table*}

\section{Experiment}
\subsection{Datasets}
We conduct main experiments on the public virtual try-on benchmark dataset VITON~\cite{viton}, which contains 14,221 data pairs for training and 2,032 data pairs for testing. 
Also, we carry out the comparative experiments under the higher resolution on the VITON-HD~\cite{viton-hd} dataset, which contains 11,647 data pairs for training and 2,032 data pairs for testing.

\subsection{Implementation Details}
We train three modules of SCW-VTON independently. For the dual-path clothing warping module and the semantic layout estimation module, they are both trained for 30 epochs and optimized by Adam~\cite{adam} with $\beta_1$=0.5 and $\beta_2$=0.999. The learning rate is fixed at 0.0001 in the first half of training and then linearly decays to zero in the remaining steps. We set the hyper-parameters as $\lambda_{per}=5$, $w_0=w_1=w_2=w_6=1$, and $w_3=w_4=w_5=3$.
For the limb-weighted try-on synthesis module, following \cite{dci-vton}, we use AdamW \cite{adamW} optimizer with the learning rate of 0.00001 to train this module for 50 epochs, and the hyper-parameters $\lambda_{vgg}$ is set to 0.0001.

\subsection{Qualitative Results}
We first compare our SCW-VTON with existing virtual try-on methods ACGPN~\cite{acgpn}, SDAFN~\cite{sdafn}, RMGN~\cite{rmgn}, and DOC-VTON~\cite{OccluMix} on the VITON~\cite{viton} dataset qualitatively. 
In Figure \ref{fig:quality}, we divide the comparison into two groups, where the left group focuses on the clothing and the right group focuses on the person's limbs. 
On the one hand, most baseline methods struggle to handle clothing with dense and complex textures.
Take the first row of the left group in Figure \ref{fig:quality} as an example, the logo after deformation presents a large range of distortions in the results of RMGN~\cite{rmgn} and DOC-VTON~\cite{OccluMix}, while the results obtained by ACGPN~\cite{acgpn} and SDAFN~\cite{sdafn} appear blurry.
In comparison, benefiting from the proposed shape constraints on clothing warping, the global shape and pose consistency of clothing is warranted in our results, leading to remarkable realism.
On the other hand, it is also challenging to fit in-shop clothing into a person image with a complex pose, which mainly involves limb occlusion and rotation. The right group of Figure \ref{fig:quality} showcases the inferior performance of baseline methods in these difficult cases. 
For instance, in the first row of the right group, ACGPN~\cite{acgpn} and SDAFN~\cite{sdafn} generate distorted and unnatural limbs, while RMGN~\cite{rmgn} and DOC-VTON~\cite{OccluMix} disrupt the continuity of limb textures.
Compared to baseline methods, we achieve more reasonable effects, which are attributed to the consistent clothing shape and realistic limb textures acquired by our method.
Qualitative experiments on the VITON-HD~\cite{viton-hd} dataset are conducted in Figure \ref{fig:quality-hd}, where the results of HR-VTON~\cite{hr-vton}, SAL-VTON~\cite{sal-vton}, DCI-VTON~\cite{dci-vton}, StableVITON~\cite{stableviton}, and our SCW-VTON are represented.
Similarly, most baseline methods struggle to align clothing with the person's body while maintaining the integrity of the clothing design. 
For example, in the first row on the left side of Figure \ref{fig:quality-hd}, most baseline methods fail to accurately estimate the correct position of the waistband, and even lose the texture information of the waistband after the clothing deformation.
In contrast, our method can preserve the waistband intact and deform it to the appropriate position according to the person's posture.
These results validate the robust performance of our method even with an increased resolution.
Additional qualitative experimental results are provided in the supplementary material.

\subsection{Quantitative Results}
We conduct quantitative experiments both in paired data setting and unpaired data setting, denoting that a person wears original clothing and new clothing after try-on, respectively. We use Structural Similarity (SSIM)~\cite{ssim} and Peak Signal to Noise Ratio (PSNR)~\cite{psnr} as metrics for the paired data setting and Fréchet Inception Distance (FID)~\cite{fid} for the unpaired data setting.
To perform a more comprehensive evaluation, we compute metrics both for the process of clothing warping and overall try-on. 
Table~\ref{tab:quantitative} summarizes the quantitative results of our method and baseline methods on two datasets, which indicates that our SCW-VTON outperforms all the baseline methods in SSIM, PSNR, and FID, both in terms of the warped clothing and try-on result. 

\begin{figure*}[!t]
    \centering
    \includegraphics[width=0.97\linewidth]{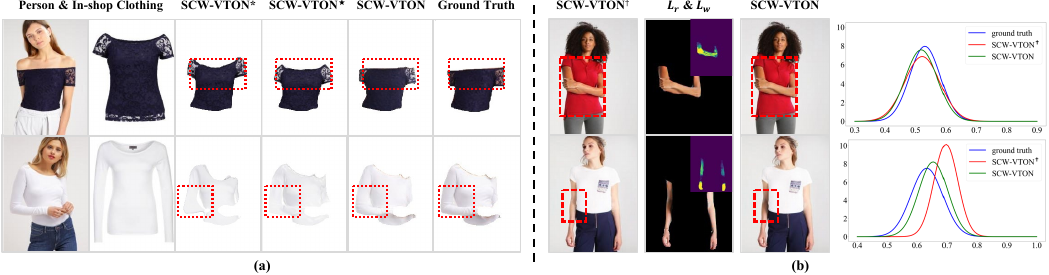}
    \vspace{-8pt}
   \caption{Qualitative results of ablation studies. (a) The variant SCW-VTON$^{\ast}$ removes shape-guided cross-attention blocks, while SCW-VTON$^{\star}$ only removes the co-training strategy. (b) SCW-VTON$^{\dagger}$ produces try-on results without the limb reconstruction network. The curve graphs on the right represent the pixel value distribution of corresponding samples in the limb regions.}
\label{fig:ab}
\end{figure*}

\begin{table*}[!t]\small
\centering
\caption{Quantitative results of ablation studies on the VITON~\cite{viton} dataset.}
\vspace{-5pt}
\setlength{\tabcolsep}{13pt}{
\scalebox{1}{
\begin{tabular}{l|c|cc|ccc}
\Xhline{2\arrayrulewidth}
\multirow{2}{*}{Method}      & \multirow{2}{*}{Config}                & \multicolumn{2}{c|}{Warped Clothing} & \multicolumn{3}{c}{Try-on Results}                \\ \cline{3-7} 
\multicolumn{1}{c|}{}        &                                        & SSIM (↑)          & PSNR (↑)         & FID (↓)       & SSIM (↑)        & PSNR (↑)        \\ \hline \hline
SCW-VTON$^{\ast}$            & w/o global shape constraints           & 0.8423            & 19.95            & 10.16          & 0.8639          & 25.35           \\
SCW-VTON$^{\star}$           & w/o co-training strategy               & 0.8792            & 22.18            & 9.44          & 0.8723          & 26.31           \\
SCW-VTON$^{\dagger}$         & w/o limb reconstruction network        & -                 & -                & 9.24          & 0.8740          & 25.75           \\ \hline
SCW-VTON (ours)              & full model                       & \textbf{0.8839}   & \textbf{23.31}   & \textbf{8.89} & \textbf{0.8897} & \textbf{26.78}  \\ \hline
\Xhline{2\arrayrulewidth}
\end{tabular}}
\label{tab:ablation}}
\vspace{5pt}
\end{table*} 

\begin{figure}[!t]
    \includegraphics[width=1.0\linewidth]{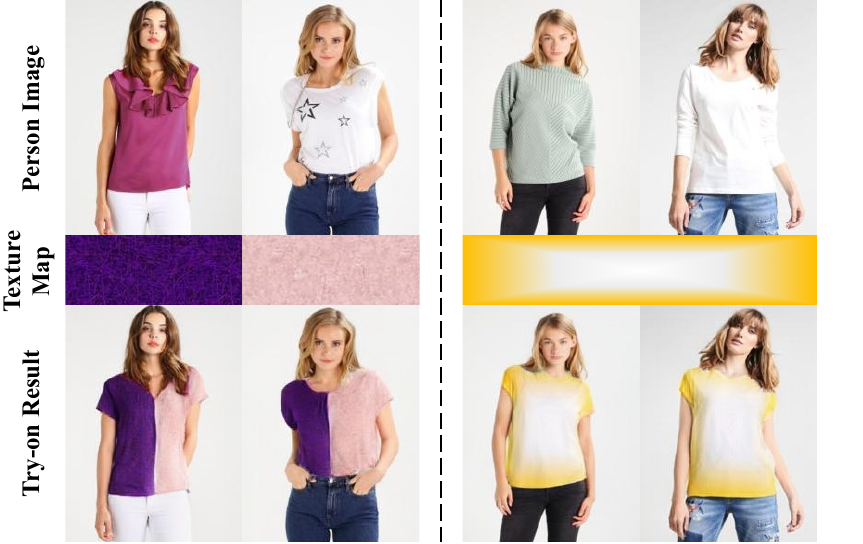}
   \caption{Examples of try-on results by our SCW-VTON based on the pure texture map input.}
\label{fig:bg}
\end{figure}

\subsection{User Study}
We conduct a user study with 30 recruited volunteers to evaluate the visual effect of our method, where SCW-VTON is compared with each baseline method in an A/B manner.
Specifically, 200 in-shop clothing images and 200 person images are randomly selected from the testing set in each comparison, which are used to output the try-on results by SCW-VTON and another baseline method to form 200 pairwise result groups.
The volunteers are asked to consider the rationality of two try-on results from each group carefully and choose the better one.
The results are summarized in the last column Table \ref{tab:quantitative}, it is evident that our SCW-VTON always provides a better visual experience in each A/B comparison.

\subsection{Ablation Studies}
\noindent\textbf{Global Shape Constraints.} 
To assess the impact of the cross-attention features $F_{att}$ as global shape constraints, we design a variant SCW-VTON$^{\ast}$ that removes shape-guided cross-attention blocks, where the flow decoder only takes the shared features $F_{style}$ as the input.
We first compare the variant and our full model in Figure \ref{fig:ab} (a). It is evident that SCW-VTON$^{\ast}$ is difficult to align well with the person's body, especially at the edge of clothing, while SCW-VTON estimates the clothing shape more accurately, and effectively addresses the problem of drastic deformation occurring in the variant.
Quantitative results of SCW-VTON and SCW-VTON$^{\ast}$ are reported in Table \ref{tab:ablation}, which shows that SCW-VTON outperforms the variant across all metrics.
The supplementary material provides more analysis and results on the global shape constraints.

\noindent\textbf{Co-training.} 
We design a variant SCW-VTON$^{\star}$ to verify the effect of the co-training in the dual-path clothing warping module.
For this variant, we only remove the co-training strategy implemented in the shape decoder and the flow decoder, while keeping all other components consistent with the full model.
SCW-VTON and SCW-VTON$^{\star}$ are quantitatively compared in Table \ref{tab:ablation}, while Figure \ref{fig:ab} (a) illustrates their differences in visual effects. It can be seen that although some misalignment issues in SCW-VTON$^{\ast}$ are alleviated in SCW-VTON$^{\star}$, SCW-VTON$^{\star}$ lacks realism in detailed regions compared to SCW-VTON.
The supplementary material provides more analysis and results on the co-training strategy.

\noindent\textbf{Limb Reconstruction Network.} 
Finally, we propose a variant SCW-VTON$^{\dagger}$ to evaluate the role of the limb reconstruction network in the process of try-on synthesis.
We ablate the autoencoder that outputs the reconstructed limb map $L_r$, so the final try-on result is generated only based on the combined representation $I_{com}$ without additional limb texture references.
We compute the quantitative metrics for SCW-VTON$^{\dagger}$, which is also shown in Table \ref{tab:ablation}. 
In addition, the qualitative comparison between this variant and SCW-VTON is presented in Figure \ref{fig:ab} (b), which illustrates that SCW-VTON can effectively solve the problems of performance degradation and distortions in SCW-VTON$^{\dagger}$, and make the distribution of reconstructed limb textures more consistent with the ground truth.

\subsection{Potential Application}
Besides the classic virtual try-on, we also try to explore other applications of our method. 
By introducing extra global shape constraints, our method is liberated from the restriction of generating try-on results solely based on specific input clothing shapes.
For example, as shown in Figure \ref{fig:bg}, we can seamlessly transfer textures from a shapeless texture map to the person's body while preserving the original distribution. 
This versatility highlights that our method extends beyond conventional virtual try-on applications, potentially sparking innovative ideas for novel computer vision tasks in the fashion and clothing domain.

\section{Conclusion}
In this paper, we propose SCW-VTON, a novel shape-guided clothing warping method for virtual try-on.
Based on a dual-path clothing warping module, our method incorporates additional global shape constraints on clothing deformation, resulting in realistic warped clothing that conforms accurately to the person's body.
Besides, we design a limb reconstruction network based on masked image modeling to learn the compact latent limb representations and generate additional limb textures to refine the details in these regions through adaptive weighting.
With the implementation of SCW-VTON, we produce try-on results with improved clothing shape consistency and precise control over details. 
Extensive experiments are conducted to demonstrate the superiority of our SCW-VTON over existing state-of-the-art methods.

\begin{acks}
This work was supported by the National Natural Science Foundation of China (No. 62072141).
\end{acks}

\bibliographystyle{ACM-Reference-Format}
\bibliography{2_references}

\appendix
\clearpage

\twocolumn[
\begin{center}
    \fontsize{14pt}{30pt}\textbf{{Supplementary Material: Shape-Guided Clothing Warping for Virtual Try-On}}
\end{center}
\vspace{20pt}
]

\noindent This document presents the supplementary material omitted from the main paper.
In Section \ref{sec1}, we provide a more detailed explanation of the semantic-replacement strategy.
In Section \ref{sec2}, we provide more qualitative results.
In Section \ref{sec3}, we present additional ablation studies on the global shape constraints and the co-training strategy.

\section{Semantic-replacement Strategy}
\label{sec1}
\begin{figure}[!t]
    \includegraphics[width=1.0\linewidth]{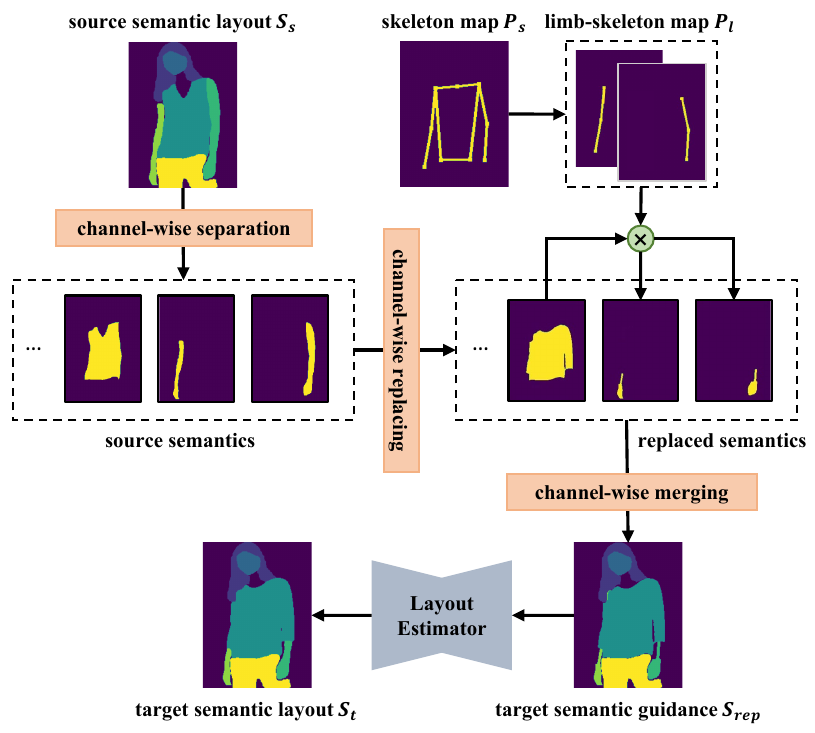}
    \caption{The schematic of the semantic-replacement layout estimation module. Based on the semantic-replacement strategy, this module produces the target semantic layout $S_{t}$ that describes the person wearing new clothing.}
\label{fig:sup_module2}
\end{figure}

Before acquiring the target semantic layout $S_t$, we introduce target semantic guidance $S_{rep}$ based on a semantic-replacement strategy, which removes the original clothing semantics while ensuring the continuity and pose consistency of the new semantics.
As shown in Figure \ref{fig:sup_module2}, the source semantic layout $S_{s}$ is first separated into a multi-channel binary parsing map, where each channel corresponds to clothing or a part of the person's body.
Except for clothing and limbs, other contents of the person image should be preserved after try-on, so we only replace the clothing channel and limb channels of the source semantic layout $S_{s}$ to get the target semantic guidance $S_{rep}$. Note that the contour of limbs reflects the shape of the clothing, so they also need to be replaced.
For the clothing channel, we replace its content with the warped clothing mask $M_{w}$, which has been aligned with the person's body in the dual-path clothing warping module.
For the limb channels, we first extract the corresponding parts from the skeleton map $P_s$ according to 6 key points of limb regions as the limb-skeleton map $P_l$. Then we mask $P_l$ with $1-M_{w}$ to discard the regions that conflict with the warped clothing due to the higher priority of clothing semantics, which is utilized to replace the original contents in the limb channels (with hand regions preserved). After the replacement, we recalibrate the last channel (representing the background region) of $S^{i}_{rep}$ based on other channels.
The overall semantic-replacement strategy can be presented as follows:
\begin{eqnarray}
    S^{i}_{rep}=\left\{
    \begin{array}{rlc}
        & S^{i}_{s}           & {0\leq i\leq 2}\\
        & M_{w}               & {i=3}\\
        & P_l\odot(1-M_{w})   & {4\leq i\leq 5}\\
        & 1-\mathcal{T}(\sum_{k=0}^{5}S^{k}_{rep})  & i=6
    \end{array} \right.,
\end{eqnarray}
where $i \in \{0,1,...,6\}$ denotes the channel index of $S_{s}$ and $S_{rep}$, $\mathcal{T}(\cdot)$ is a truncated function that ensures the output value falls within the range of zero and one, and $\odot$ is the element-wise multiplication. Then, the content of each $S^{i}_{rep}$ is channel-wise merged again.

\begin{figure*}
    \centering
    \includegraphics[width=0.95\linewidth]{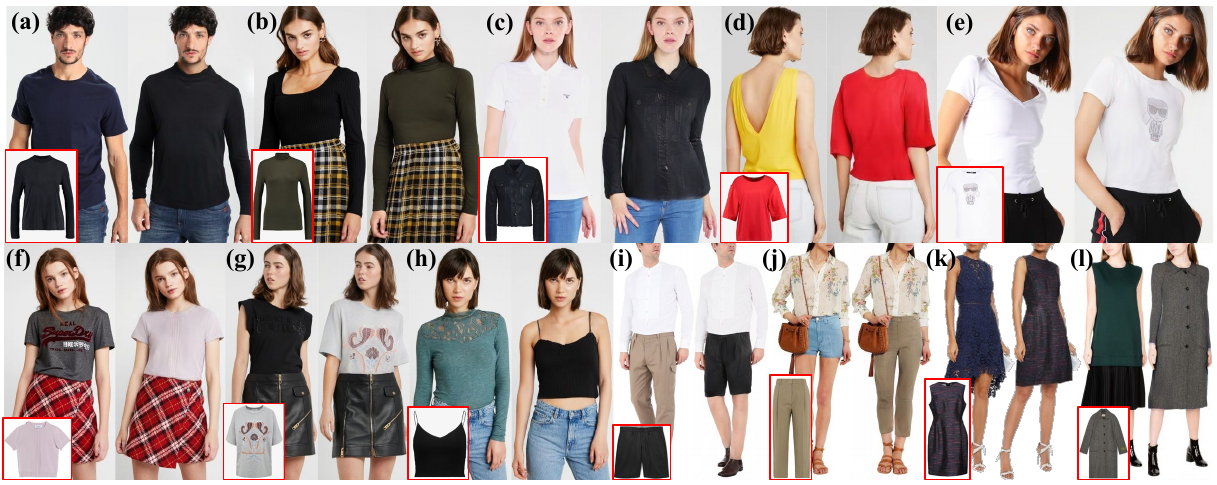}
    \caption{More specific samples of our method.}
    \label{fig:results1}
\end{figure*}

\begin{figure}[!t]
    \includegraphics[width=0.98\linewidth]{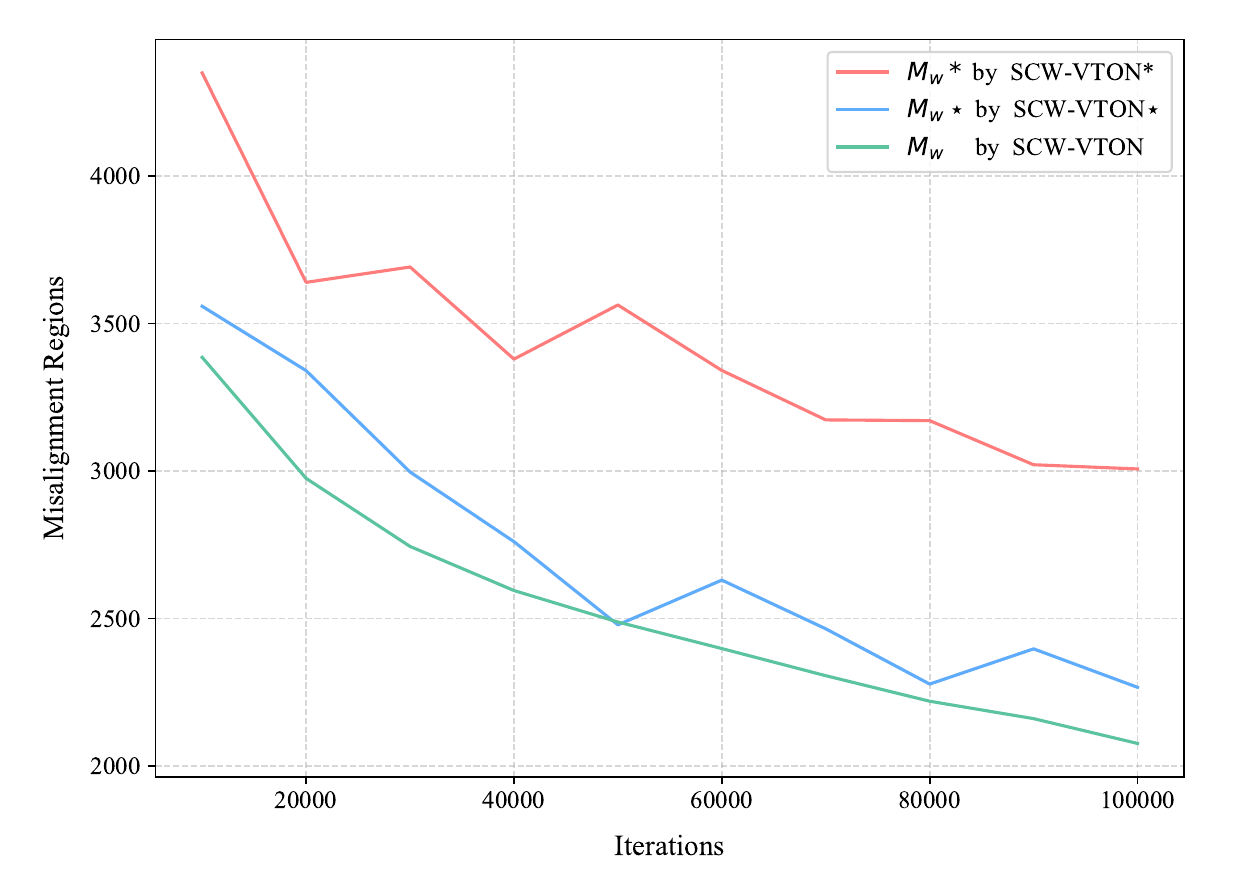}
    \caption{The number of pixels in the misalignment regions caused by $M^{\ast}_w$, $M^{\star}_w$, and $M_w$ as the training iteration increases.}
\label{fig:vs}
\end{figure}

Subsequently, we employ a UNet~\cite{unet} model as the semantic layout estimator, which predicts the target semantic layout $S_{t}$ by inputting the target semantic guidance $S_{rep}$.
Since the pose information and the semantics of warped clothing have been integrated into the input beforehand, the estimator can more effectively discern the generation locations of target semantics based on the contents after replacement, which provides more accurate structural guidance for the subsequent try-on synthesis.

\section{More Qualitative Results}
\label{sec2}
\begin{figure*}[!t]
    \centering
    \includegraphics[width=0.9\linewidth]{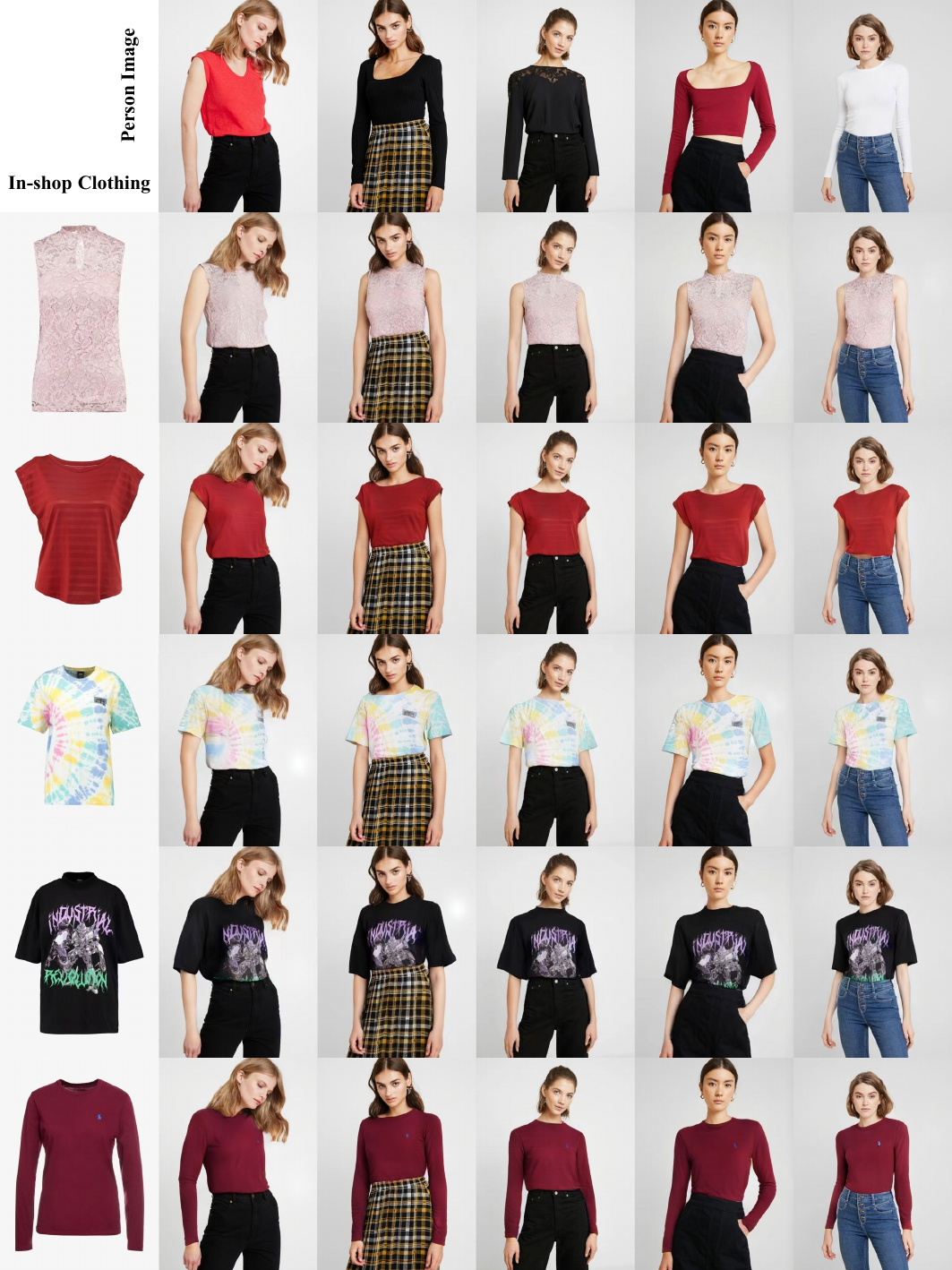}
   \caption{In-shop clothing with different styles and person images with different poses are selected from the testing set of the VITON-HD~\cite{viton-hd} dataset to form multiple input groups and generate diverse try-on results.}
\label{fig:robust2}
\end{figure*}

We conduct additional qualitative experiments to validate the diversity of the results generated by our method. 
First, we select in-shop clothing with different styles and person images with different poses from the testing set of the VITON-HD~\cite{viton-hd} dataset, which are combined into pairs to form multiple input groups, producing various try-on results.
These results are depicted in Figure \ref{fig:robust2}, showcasing the robustness and generalization capability of our SCW-VTON, which is not contingent upon specific paired data.
Furthermore, as shown in Figure \ref{fig:results1}, we provide more specific samples of (a) male model, (b) turtleneck, (c) jacket, (d) back pose, (e) side pose, (f) children's clothing, (g) big logo, (h) crop top, (i) shorts, (j) pants, (k) dress, and (l) coat.

\begin{figure*}[!t]
    \includegraphics[width=1.0\linewidth]{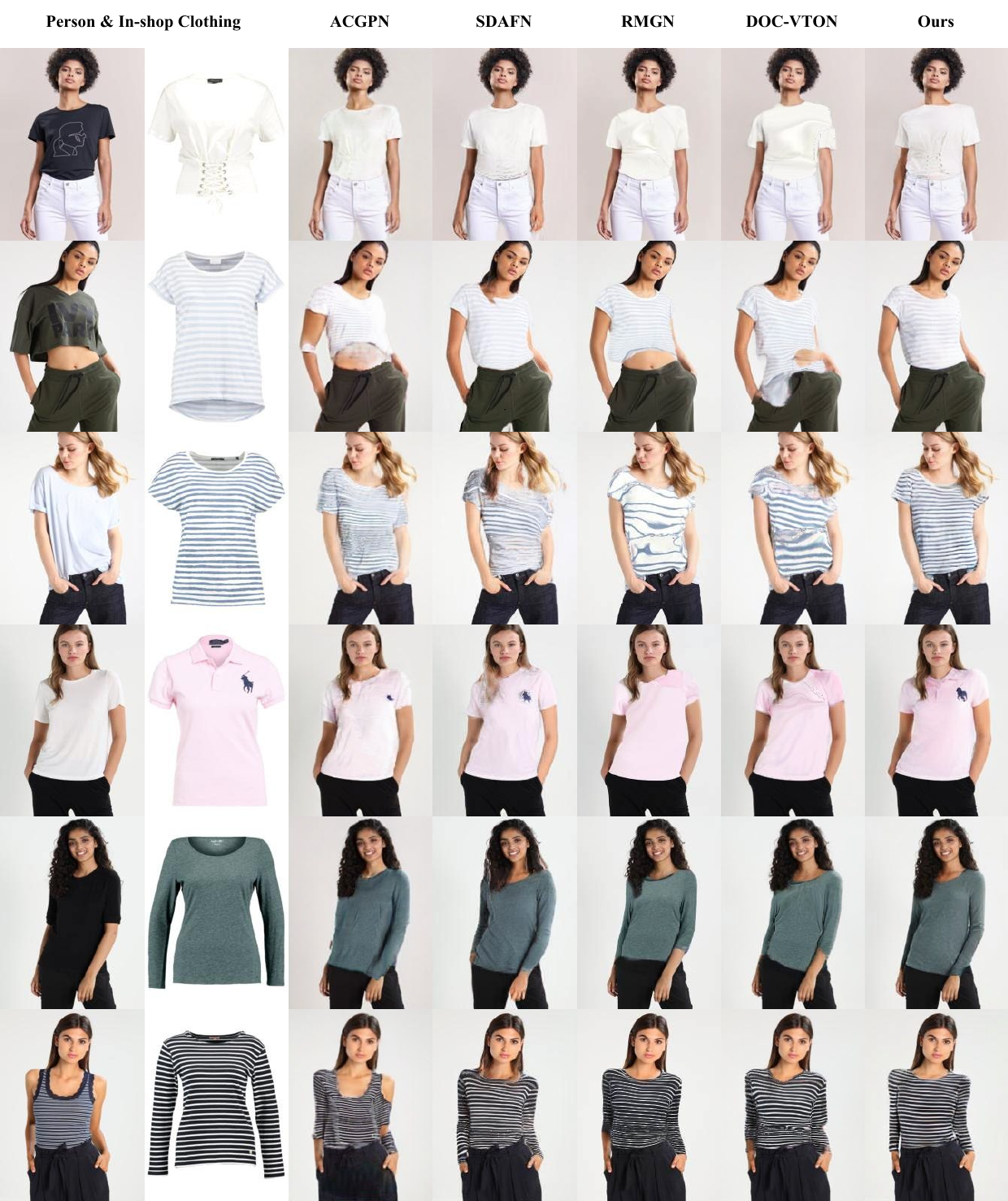}
    \caption{Qualitative results of our SCW-VTON and baseline methods on the VITON~\cite{viton} dataset.}
\label{fig:sup_quality}
\end{figure*}

\begin{figure*}[!t]
    \includegraphics[width=1.0\linewidth]{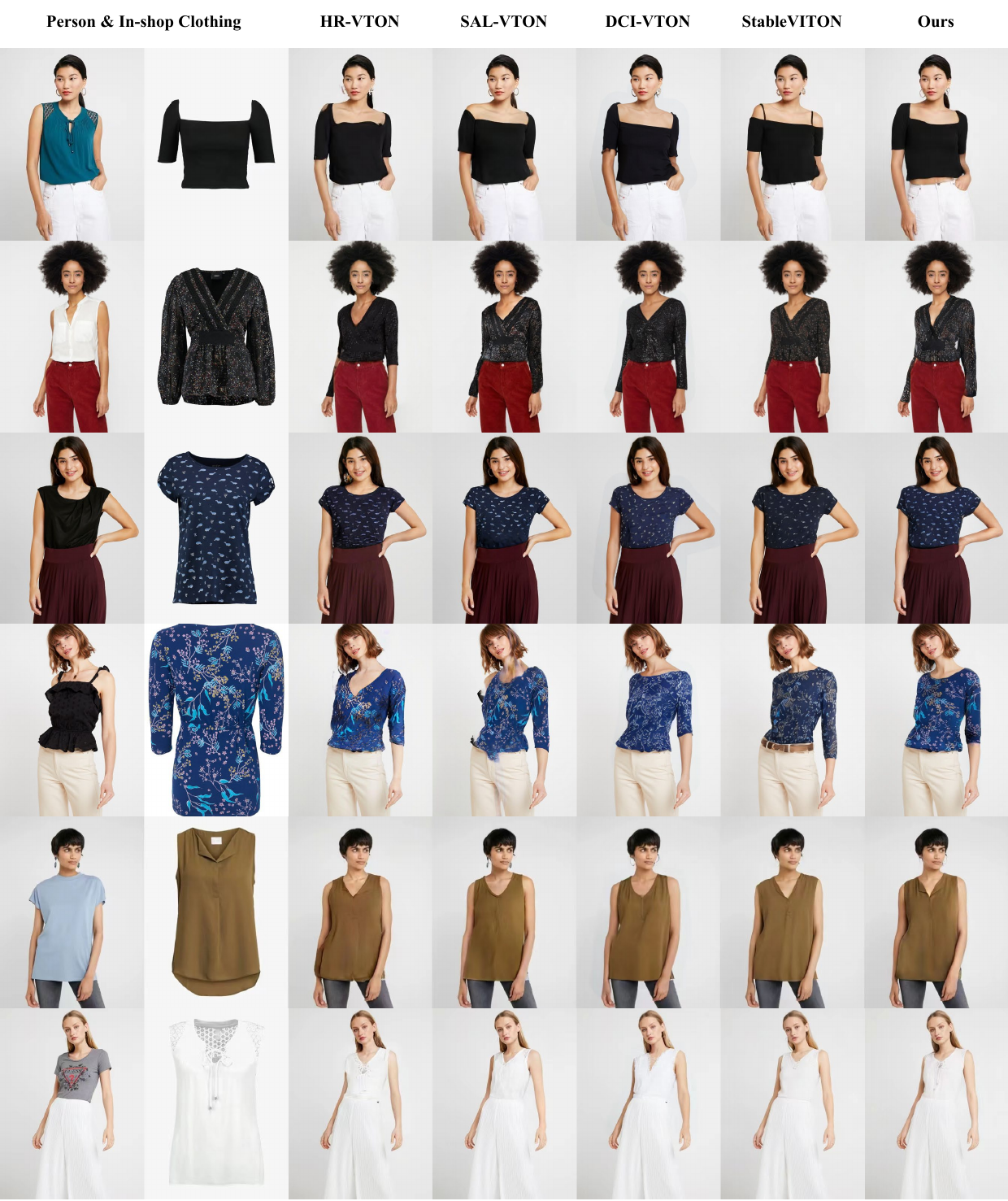}
    \caption{Qualitative results of our SCW-VTON and baseline methods on the VITON-HD~\cite{viton-hd} dataset.}
\label{fig:sup_quality_hd}
\end{figure*}

Additionally, Figure \ref{fig:sup_quality} provides more qualitative comparison results of ACGPN~\cite{acgpn}, SDAFN~\cite{sdafn}, RMGN~\cite{rmgn}, DOC-VTON~\cite{OccluMix}, and SCW-VTON on the VITON~\cite{viton} dataset. 
Figure \ref{fig:sup_quality_hd} exhibits further comparison results of HR-VTON~\cite{hr-vton}, SAL-VTON~\cite{sal-vton}, DCI-VTON~\cite{dci-vton}, StableVITON~\cite{stableviton}, and SCW-VTON on the VITON-HD~\cite{viton-hd} dataset.

\section{More Ablation Studies}
\label{sec3}
We conduct additional experiments to demonstrate the necessity of incorporating extra global shape constraints and the co-training strategy.
We adopt the same setup as in the main paper, defining SCW-VTON$^{\ast}$ as a variant of SCW-VTON that ablates shape-guided cross-attention blocks, while SCW-VTON$^{\star}$ is another variant that only ablates the co-training strategy.
Specifically, we compare the ability of SCW-VTON, SCW-VTON$^{\ast}$, and SCW-VTON$^{\star}$ to capture the shape characteristics of clothing after deformation with increasing training iterations.
To distinguish from the warped clothing mask $M_w$ obtained from our SCW-VTON, we denote the output mask by SCW-VTON$^{\ast}$ as $M^{\ast}_w$, and the output mask by SCW-VTON$^{\star}$ as $M^{\star}_w$. 
We respectively calculate the misalignment regions of $M^{\ast}_w$, $M^{\star}_w$, and $M_w$ with the ground truth $M_{gt}$ (i.e., the actual clothing regions in the person image) as the number of training iterations increases, and compare these misalignment regions in Figure \ref{fig:vs}.
The results illustrate that $M^{\ast}_w$ generates a considerable number of misalignment regions, and it is evident that the appearance flow predicted by SCW-VTON$^{\ast}$ is unstable.
By incorporating shape-guided cross-attention blocks to provide global shape constraints in SCW-VTON$^{\star}$, $M^{\star}_w$ exhibits significantly less misalignment and better matches the person's body.
Moreover, with the implementation of the co-training strategy in our full model SCW-VTON, there is further improvement in both accuracy and stability.
In summary, these results demonstrate the effectiveness of the shape-guided cross-attention blocks in providing global shape constraints for estimating the clothing shape aligned with the person's body accurately and robustly, along with the beneficial impact of co-training between the two paths on the weight update of the flow decoder.

\end{document}